\documentclass{article}
\usepackage{graphicx} 
\usepackage{authblk}
\usepackage[margin=1in]{geometry}
\usepackage{hyperref}
\usepackage[numbers]{natbib}
\usepackage{array}
\usepackage{setspace}
\usepackage{amsmath}
\usepackage{booktabs}

\title{Detection of Disease on Nasal Breath Sound by New Lightweight Architecture: Using COVID-19 as An Example}

\author[1,2,*]{Jiayuan She}
\author[3,*]{Lin Shi}
\author[4,*]{Peiqi Li}
\author[1]{Ziling Dong}
\author[1,5]{Renxing Li}
\author[4]{Shengkai Li}
\author[3]{Liping Gu}
\author[6]{Zhao Tong}
\author[1]{Zhuochang Yang}
\author[7,$\dag$]{Yajia Ji}
\author[2,$\dag$]{Liang Feng}
\author[1,$\dag$]{Jiangang Chen}

\affil[1]{\small Shanghai Key Laboratory of Multidimensional Information Processing, East China Normal University, Shanghai, China \\ \texttt{Lucas\_SJY@outlook.com (Jiayuan She); mzn3222022@gmail.com (Ziling Dong); aa1773894229@163.com (Renxing Li); yafrank@amazon.com (Zhuochang Yang); \newline jgchen@cee.ecnu.edu.cn (Jiangang Chen)}}
\affil[2]{\small School of Artificial Intelligence \& Advanced Computing, Xi'an Jiaotong-Liverpool University (Taicang Campus), Taicang, China}
\affil[3]{\small Department of Ultrasound in Medicine, Shanghai Sixth People's Hospital Affiliated to Shanghai Jiaotong University School of Medicine, Shanghai, China \newline \texttt{shilin\_love@163.com (Shi Lin); Guliping666@126.com (Liping Gu); fenliangg@163.com (Liang Feng)}}
\affil[4]{\small School of Mathematics \& Physics, Xi'an Jiaotong-Liverpool University, Suzhou, China \newline \texttt{LPQ\_0619@outlook.com (Peiqi Li); Shengkai.Li23@student.xjtlu.edu.cn (Shengkai Li)}}
\affil[5]{\small School of Advanced Technology, Xi'an Jiaotong-Liverpool University, Suzhou, China}
\affil[6]{\small Institute of Information Engineering, Chinese Academy of Science, Beijing, China \newline \texttt{tongzhao@iie.ac.cn (Zhao Tong)}}
\affil[7]{\small Shuguang Hospital Affiliated to Shanghai University of Traditional Chinese Medicine, Shanghai, China \newline \texttt{2373602874@qq.com (Yejie Ji)}}
\affil[*]{Jiayuan She, Shi Lin, and Peiqi Li are the co-first authors.}
\affil[$\dag$]{Jiangang Chen, Liang Feng, and Yajie Ji are the corresponding authors}

\begin{document}
\date{}
\maketitle

\begin{abstract}
    \textit{Background}. Infectious diseases, particularly COVID-19, continue to be a significant global health issue. Although many countries have reduced or stopped large-scale testing measures, the detection of such diseases remains a propriety. \textit{Objective}. This study aims to develop a novel, lightweight deep neural network  for efficient, accurate, and cost-effective detection of COVID-19 using a nasal breathing audio data collected via smartphones. \textit{Methodology}. Nasal breathing audio from 128 patients diagnosed with the Omicron variant was collected. Mel-Frequency Cepstral Coefficients (MFCCs), a widely used feature in speech and sound analysis, were employed for extracting important characteristics from the audio signals. Additional feature selection was performed using Random Forest (RF) and Principal Component Analysis (PCA) for dimensionality reduction. A Dense-ReLU-Dropout model was trained with K-fold cross-validation (K=3), and performance metrics like accuracy, precision, recall, and F1-score were used to evaluate the model. \textit{Results}. The proposed model achieved 97\% accuracy in detecting COVID-19 from nasal breathing sounds, outperforming state-of-the-art methods such as those by \cite{lella2021literature} and \cite{abayomi2022detection}. Our Dense-ReLU-Dropout model, using RF and PCA for feature selection, achieves high accuracy with greater computational efficiency compared to existing methods that require more complex models or larger datasets. \textit{Conclusion}. The findings suggest that the proposed method holds significant potential for clinical implementation, advancing smartphone-based diagnostics in infectious diseases. The Dense-ReLU-Dropout model, combined with innovative feature processing techniques, offers a promising approach for efficient and accurate COVID-19 detection, showcasing the capabilities of mobile device-based diagnostics
\end{abstract}


\section{Introduction}
The global outbreak of COVID-19 in 2019 has posed unprecedented challenges to public health systems, with emerging variants such as Omicron exacerbating the crisis due to heightened transmissibility and severe respiratory complications\cite{richards2020impact}. Despite advancements in diagnostic tools like RT-PCR and rapid antigen tests, widespread implementation remains hindered by cost, accessibility, and logistical barriers, particularly in resource-limited regions\cite{usman2020possibility}. This gap underscores the urgent need for non-invasive, scalable, and cost-effective diagnostic alternatives to curb transmission and enable timely interventions.

Recent advancements in digital health technologies have highlighted the potential of acoustic analysis for disease detection. Respiratory infections, including COVID-19, often alter vocal fold dynamics, breathing patterns, and sound production, offering a unique opportunity to leverage audio signals as diagnostic biomarkers\cite{deshmukh2021interpreting}. Speech and cough analysis have been widely explored, with studies demonstrating the feasibility of machine learning models in detecting COVID-19 through vocalizations (e.g., vowel articulation, cough sounds). For instance, deep learning models analyzing cough recordings have achieved validation accuracies of 67–83\%\cite{al2021detection}, while vocal fold vibration analysis via vowel vocalization has yielded ~80\% accuracy\cite{jeleniewska2022isolated}. However, these approaches often require complex speech tasks or extensive computational resources, limiting their practicality for real-world deployment\cite{mari2024voice}.

A critical gap in existing research lies in the underutilization of nasal breathing sounds—a passive, non-invasive signal that directly reflects upper respiratory tract physiology. Unlike speech or cough sounds, nasal breathing is effortless, making it ideal for rapid screening in diverse populations, including asymptomatic individuals\cite{miranda2019comparative}. While studies have explored oral breathing and cough acoustics for COVID-19 detection\cite{imran2020ai4covid,khanzada2021challenges,laguarta2020covid}, nasal breathing remains underexamined despite its clinical relevance. Existing methodologies also face challenges such as low accuracy (<85\% in some studies\cite{xiong2022reliability}[), reliance on high-dimensional datasets, and insufficient feature optimization, which can introduce noise and overfitting \cite{despotovic2021detection}.

To address these limitations, this study proposes a lightweight deep neural network (DNN) framework optimized for nasal breathing sound analysis. By integrating Mel-Frequency Cepstral Coefficients (MFCCs) with advanced feature selection techniques—Random Forest (RF) and Principal Component Analysis (PCA)—we aim to reduce computational complexity while enhancing diagnostic accuracy\cite{abayomi2022detection,benmalek2023automatic}. Our approach leverages smartphone-recorded nasal breathing sounds from 128 Omicron patients, focusing on key acoustic features such as fundamental frequency, sound pressure level, and MFCCs\cite{bagad2020cough,costantini2022deep}. Through systematic dimensionality reduction and 3-fold cross-validation, we evaluate the robustness of our model against state-of-the-art methods, demonstrating its potential for scalable, real-world clinical applications.

The primary contributions of this work are threefold:
\begin{enumerate}
    \item \textbf{Novel Data Source}: First large-scale exploration of nasal breathing sounds for COVID-19 detection, capturing upper airway pathophysiology.
    \item \textbf{Lightweight Architecture}: A computationally efficient Dense-ReLU-Dropout DNN model optimized for mobile deployment.
    \item \textbf{Feature Engineering}: Integration of RF and PCA to enhance model generalizability, achieving 97\% accuracy with minimal feature redundancy.
\end{enumerate}

The remainder of this paper is organized as follows: \hyperref[sec:methods]{Section.\ref{sec:methods}} details the dataset collection, preprocessing, and feature extraction methodologies. Section 3 describes the lightweight DNN architecture and experimental setup. Results and comparative analyses are presented in Section 4, followed by a discussion of clinical implications and limitations in Section 5. Finally, Section 6 concludes the study and outlines future research directions.

\section{Materials and Methods}
\label{sec:methods}

\subsection{Experiment Design}
\begin{figure}[t]
    \centering
    \includegraphics[width=1.0\linewidth]{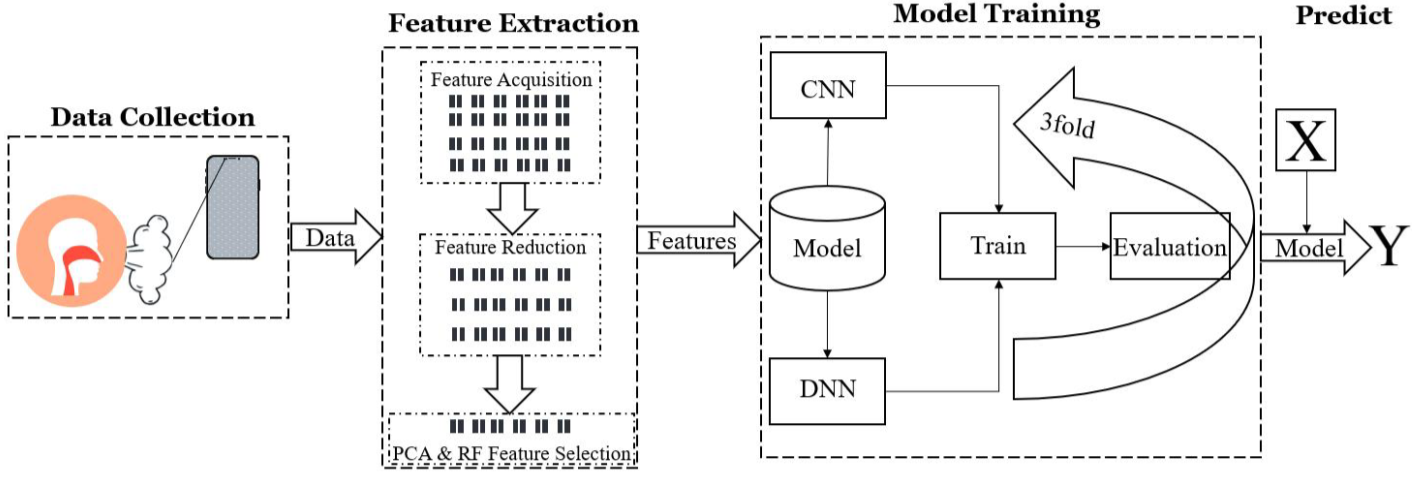}
    \caption{The general flow of the study consists of four parts: Data Collection, Feature Extraction, Model Training, Prediction.}
    \label{fig:flow_chart}
\end{figure}
\hyperref[fig:flow_chart]{Fig.\ref{fig:flow_chart}} shows our research methodology, which is divided into four stages: data collection, feature extraction, model training, and prediction.

1. \textbf{Data Collection}: Nasal breathing data from patients is collected using a smartphone, providing real-time respiratory samples for analysis.

2. \textbf{Feature Extraction}: Crucial features are selected from raw data through operations like dimensionality reduction and secondary feature selection using PCA and RF algorithms.

3. \textbf{Model Training}: Extracted features serve as input for training and evaluating a deep learning model in a 3-fold cross-validation setup to ensure robustness.

4. \textbf{Prediction}: The trained model is used for actual predictions, offering insights into patient conditions.

\subsection{Dataset}
\subsubsection{Data Collection}
The prospective observational study was conducted over 1 year, from Mar.2021 to Feb.2022, at Shanghai Sixth People’s Hospital and Shanghai Key Laboratory of Multidimensional Information Processing, East China Normal University. Nasal breathing sounds were collected from 128 participants (67 COVID-19 positive, 61 healthy controls) using a smartphone application in a controlled clinical environment. All participants provided written informed consent, and the study protocol was approved by the institutional Ethics Committee of Shanghai Sixth People’s Hospital (Approval No. 2022-KY-050(K)). These real-time respiratory samples were then analyzed for acoustic feature extraction and disease detection.

\vspace{2pt}
\textbf{Inclusion Criteria}: Participants were included if they met the following criteria:
\begin{enumerate}
    \item Aged 18-65 years
    \item Diagnosed with COVID-19 (Omicron variant), confirmed by PCR of rapid antigen test
    \item Able to provide informed consent and participate in this study
    \item Presenting symptoms of COVID-19, such as fever, cough, or difficulty breathing, at the time of data collection
\end{enumerate}

\textbf{Exclusion Criteria}: Participants were excluded if they met any of the following criteria:
\begin{enumerate}
    \item History of severe respiratory conditions such as asthma, COPD, or pneumonia
    \item Pregnancy of breastfeeding
    \item Any known neurological disorders or hearing impairments that could affect the ability to participate in the study
    \item Use of medications that could affect respiratory function (e.g. sedatives or narcotics)
\end{enumerate}

The dataset comprises standard audio recordings of nasal breathing sounds from 67 patients diagnosed with neoplastic disease, collected from the Shanghai Sixth People’s Hospital. Each recording has an average duration of approximately six seconds. For the control group, nasal breath sounds were collected from 61 adults who tested negative for neoplastic pneumonia using the same recording procedure. Written consent was obtained from all participants, and the study was reviewed and approved by an ethics committee. Unlike other nasal breathing datasets, such as the COVID-19 dataset from Sonde Health (2020), no category in our dataset contains more than 25 individuals. This dataset provides crucial insights and supports research into the association between neoplastic disease and COVID-19 through nasal breathing sound characteristics. Additionally, it aids in the development of methods for early COVID-19 detection(see \hyperref[tab_patients]{Tab.\ref{tab_patients}} for detail). 

\begin{table}[h]
\centering
\caption{Number of patients enrolled in our study}
\label{tab:patient_count}
\begin{tabular}{|l|l|l|}
\hline
\textbf{Type} & \textbf{Positive for neoplastic disease} & \textbf{Negative for neoplastic disease} \\ \hline
Number of samples & 67 & 61 \\ \hline
\end{tabular}
\label{tab_patients}
\end{table}

\subsubsection{Data Preprocessing}
The raw nasal breathing audio signals collected via smartphones are inherently one-dimensional (1D) time-series data, representing variations in sound pressure levels over time. However, convolutional neural networks (CNNs) are traditionally designed to process two-dimensional (2D) data, such as images, where spatial hierarchies and local patterns are critical for feature extraction. To leverage the powerful pattern recognition capabilities of CNNs, we transformed the 1D audio signals into a 2D format.

This approach allows the CNN to capture local features in the data, which are essential for distinguishing COVID-19-related acoustic features from healthy controls. By reshaping the 1D audio data into 2D formats, we enable the CNN to apply convolutional filters across both time and frequency dimensions, enhancing its ability to detect subtle associated with respiratory abnormalities.

\begin{figure}[h]
    \centering
    \includegraphics[width=0.9\linewidth]{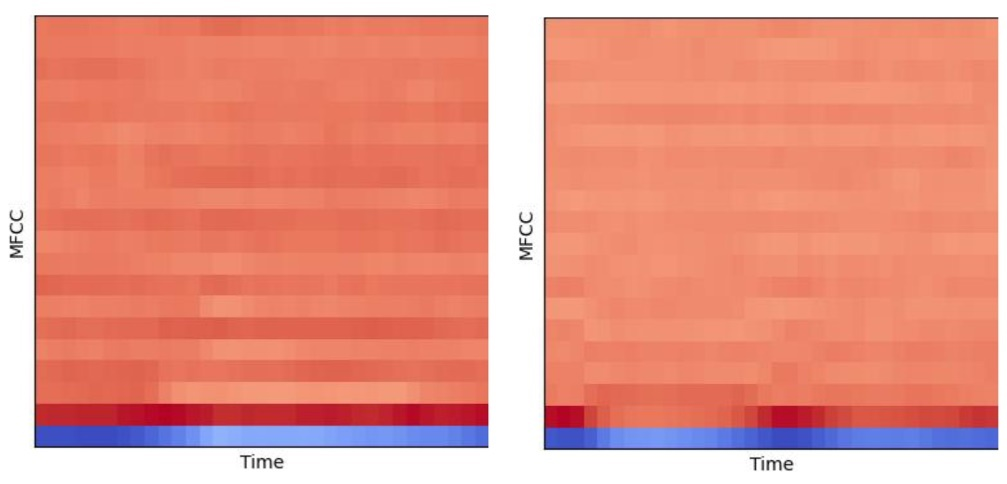}
    \caption{Mel-scale frequency cepstral coefficients map for nasal breath sound in tested adults for COVID-19 (Left: Positive; Right: Negative)}
    \label{fig_mfcc}
\end{figure}

\subsection{Statistical Analysis}
\subsubsection{Feature Selection}

This study employed nine key acoustic features to provide a comprehensive analysis of the audio signal to achieve accurate results. These characteristics include: voiced and unvoiced sounds, effective speech segments, fundamental frequency (F0), log energy, short-term energy, zero crossing rate, sound pressure level (SPL) and Mel Frequency Cepstral Coefficients (MFCCs), with detailed explanations below\cite{arora2020bolasso,zhou2016cost}:
\begin{enumerate}
\item \textbf{Voiced Sounds}: Sounds produced by vocal fold vibration, usually louder and of longer duration.
\item \textbf{Unvoiced Sounds}: Sounds that are not produced by vocal fold vibration and are usually quieter and shorter in duration.
\item Effective Speech Segments: The part of the speech signal that contains meaningful or relevant information.
\item \textbf{Fundamental Frequency (F0)}: The lowest frequency of the periodic waveform, i.e. the frequency at which the vocal folds vibrate, related to the pitch of the voice.
\item \textbf{Log energy}: A measure of the total power or loudness of a speech signal, usually calculated over a short period and then logarithmically scaled.
\item \textbf{Short-term energy}: A measure of the change in energy or amplitude over a short period, which can be used to distinguish between voiced and unvoiced speech, or to detect the presence of speech.
\item \textbf{Zero Crossing rate}: The rate at which a signal changes from positive to negative or vice versa, often used to distinguish between voiced and unvoiced speech.
\item \textbf{Sound Pressure Level (SPL)}: A logarithmic measure of the effective pressure of a sound relative to a reference value, used in acoustics to quantify sound levels.
\item \textbf{Mel Frequency Cepstral Coefficients (MFCCs)}: A set of coefficients representing the short-time power spectrum of a sound, based on a linear cosine transformation of the logarithmic power spectrum on a non-linear Mel frequency scale. They are widely used in speech and speaker recognition because of their ability to represent key features of the voice.
\end{enumerate}

Each acoustic feature provides a unique and reliable perspective on voice signal analysis. In this study, these features were used for audio data analysis, offering comprehensive acoustic insights crucial for subsequent model training. The analysis revealed significant variations in the acoustic properties of nasal breathing between individuals with different COVID-19 test results.

\hyperref[fig_mfcc]{Fig.\ref{fig_mfcc}} demonstrates that the MFCC plots of COVID-19 positive patients exhibit significantly darker tones, indicating a lower frequency component in their nasal breathing sounds compared to healthy individuals. This variation may be linked to the impact of COVID-19 infection on the sound produced.

\begin{figure}[h]
    \centering
    \includegraphics[width=0.9\linewidth]{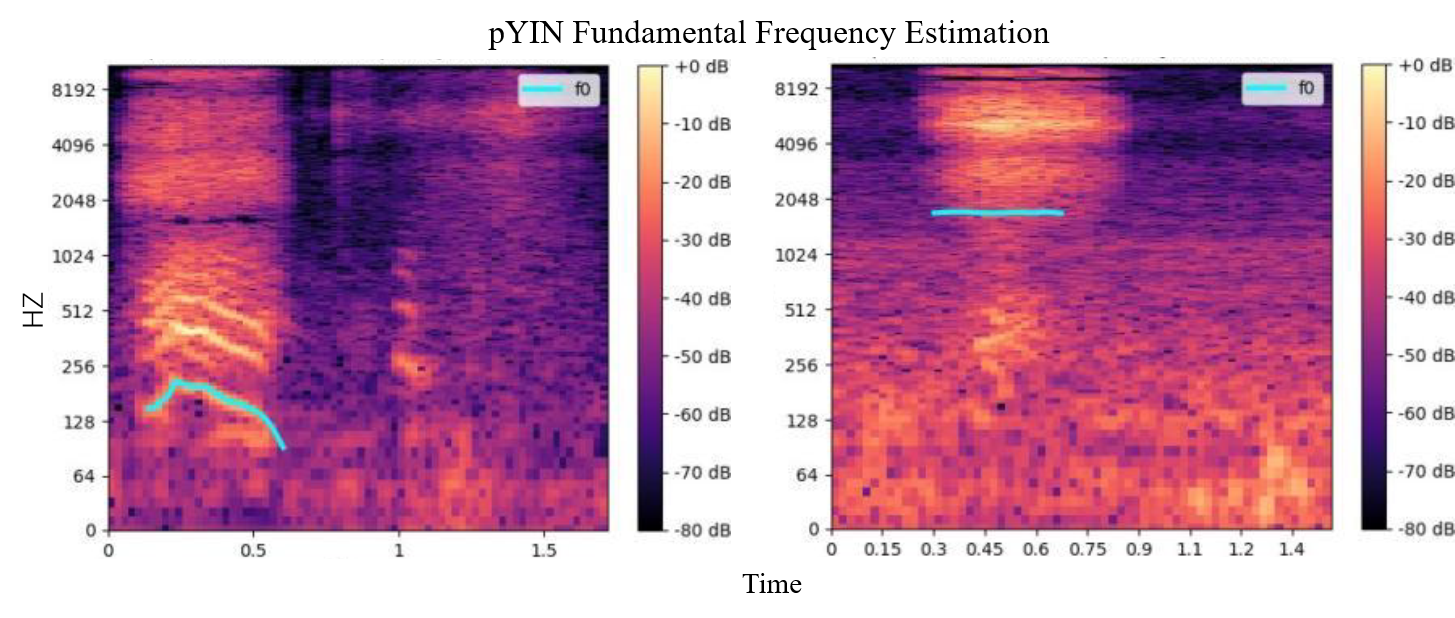}
    \caption{Fundamental frequency curve for nasal breath sound in adults who tested for COVID-19 (Left: Positive; Right: Negative)}
    \label{fig_fundamental_freq}
\end{figure}

\hyperref[fig_fundamental_freq]{Fig.\ref{fig_fundamental_freq}} shows that F0 curves of COVID-19 patients are significantly lower than those of healthy individuals, suggesting that their nasal breathing sounds are relatively low-pitched. The stable frequency observed on the right-hand side, with no lower frequency in that period, supports the notion that 2k can be the fundamental frequency.

We extract statistical information such as extreme values, mean values, standard deviations, peak values, and skewness from a range of acoustic features. These metrics collectively provide a comprehensive quantification of the key characteristics of each feature. This process offers an analysis that is more profound and informative than that provided by a two-dimensional image. Detailed statistical metrics selected can be seen in the \hyperref[tab:metrics]{Tab.\ref{tab:metrics}}:

\begin{table}[h]
\centering
\caption{Selected statistical metrics in this study}
\renewcommand{\arraystretch}{2.0}
\begin{tabular}{|l|p{6cm}|p{6cm}|}
\hline
\textbf{Metrics} & \textbf{Meanings} & \textbf{Formulas} \\ \hline
Mean value & Illustrating the central tendency & $\bar{x} = \frac{1}{n} \sum_{i=1}^{n} x_i$ \\ \hline
Standard deviation & Indicating the feature's variability & $\sigma = \sqrt{\frac{1}{n} \sum_{i=1}^{n} (x_i - \bar{x})^2}$ \\ \hline
Extreme value & The maximum sound intensity & $\max_i = \max x_i$ \\ 
 &  & $\min_i = \min x_i$ \\ \hline
Skewness & Measures the asymmetry of the distribution & Skewness$= \frac{\frac{1}{n} \sum_{i=1}^{n} (x_i - \bar{x})^3}{\left(\frac{1}{n} \sum_{i=1}^{n} (x_i - \bar{x})^2\right)^{\frac{3}{2}}}$ \\ \hline
Kurtosis & Outlining the shape of the sound feature's distribution & Kurtosis$= \frac{\frac{1}{n} \sum_{i=1}^{n} (x_i - \bar{x})^4}{\left(\frac{1}{n} \sum_{i=1}^{n} (x_i - \bar{x})^2\right)^2} - 3$ \\ \hline
Range value & Span of breath sound data & Range$= \max_i x_i - \min_i x_i$ \\ \hline
\end{tabular}
\label{tab:metrics}
\end{table}

We extract statistical information such as extreme values, mean values, standard variances, and skewness from a range of acoustic features. These metrics collectively provide a comprehensive quantification of the key characteristics of each feature. This process offers an analysis that is more profound and informative than that provided by a two-dimensional image. The extreme value outlines the sound feature's upper and lower boundaries. The mean value illustrates its central tendency. The standard variance is indicative of the feature's variability. The peak value points to the maximum sound intensity. The skewness measures the skewness of the distribution. The process of condensing data from two-dimensional images into these vital statistical metrics helps minimizes redundant information. This step contributes significantly to making the model more streamlined.

\subsubsection{Feature Dimension Reduction}
In this subsection we will introduce the methods we applied for feature dimension reduction. The model input the more redundant information involved\cite{dewi2019random,malhi2004pca,song2010feature,lella2021literature}. 

In traditional studies of COVID-19 detection utilizing acoustic features, the challenge of low accuracy often arises. To counter this, we explore critical comparisons of these acoustic features and employ a strategy designed to reduce model complexity, enhancing operational efficiency. Consider data from two-dimensional images into these vital statistical metrics helps minimize redundant information and significantly contributes to making the model more streamlined.

In recent years, dimension reduction techniques have evolved significantly, with several new methods offering promising improvements over traditional approaches. In this study, we employed Principal Component Analysis and Random Forest for the feature selection and dimensionality reduction. While newer methods such as t-SNE (t-Distributed Stochastic Neighbor Embedding) and autoencoders have gained popularity in the field, recent studies still demonstrated the effectiveness of PCA in acoustic and sound-based analysis. For example, \cite{wang2024current} show that PDCA remains a reliable method for extracting relevant features from high-dimensional audio data, especially when computational resources are constrained. Similarly, \cite{chen2024machine} found that RF-based feature selection, when combined with PCA, continues to yield high performance in diagnostic tasks using sound data, making these traditional techniques a strong choice for our study.

To maintain model efficiency, we adopted strategies including Random Forest (RF)\cite{kannagi2024data,bania2024r} and Principal Component Analysis (PCA) \cite{khadka2025geographic,jalo2024hybrid,palanikkumar2025machine} for ranking and prioritizing the extracted features. 

RF ranks and prioritizes features based on their importance scores, where the importance score for a feature $x_i$ is given by
\begin{equation}
    \text{Importance}(x_i)=\frac{1}{T}\sum_{i=1}^TI_t(x_i)
\end{equation}
where $I_t(x_i)$ indicates the importance of $x_i$ in tree $t$, and $T$ is the total number of trees.

PCA reduces dimensionality by transforming the original features into a new set of uncorrelated features (principal components) that maximize variance. The principal components are given by:
\begin{equation}
    Z=XW
\end{equation}
where $X$ is the original data matrix, and $W$ is the matrix of eigenvectors of the covariance matrix of $X$. 

Beyond the implementation of RF and PCA, we explored the correlation analysis between input features and labels. We selected the top eight features exhibiting the highest correlation coefficients to form a new subset of features. The correlation coefficient $\rho$ between feature $x$ and label $y$ is given by:
\begin{equation}
    \rho_{x,y}=\frac{Cov(x,y)}{\sigma_x\cdot\sigma_y}
\end{equation}

This approach ensures that the chosen attributes have a significant relationship with the target label, enhancing our model’s diagnostic accuracy.

\subsection{Network Architecture}
\begin{figure}[h]
    \centering
    \includegraphics[width=0.9\linewidth]{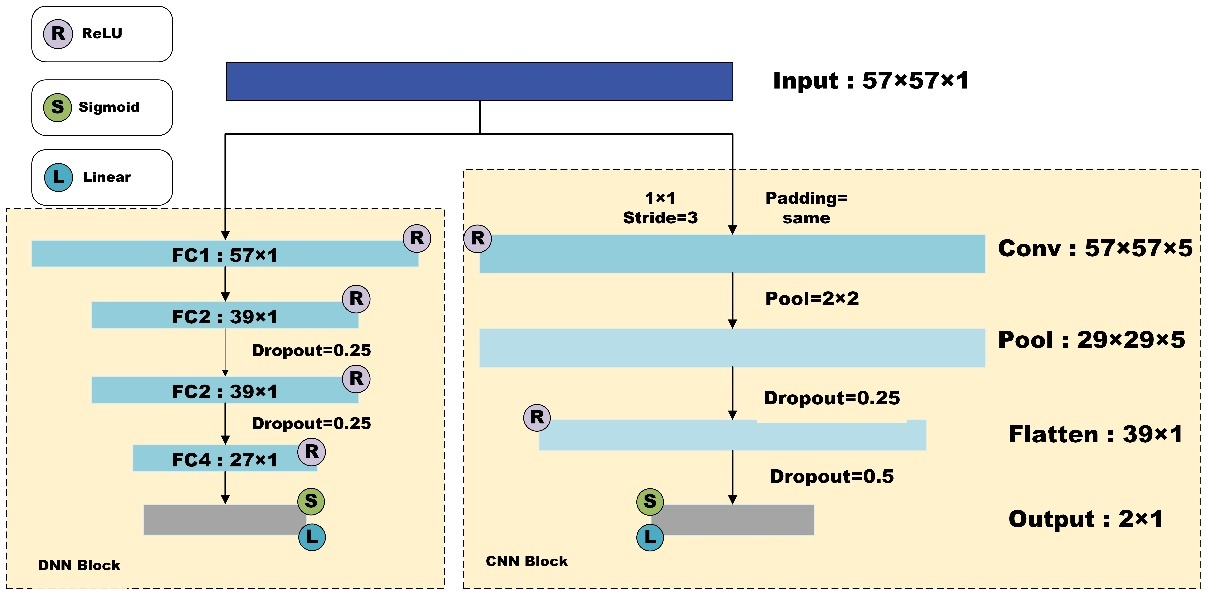}
    \caption{Network architecture of our study. ReLU: Rectified linear unit. Sigmoid: Sigmoid activation function with ‘S’ shape. Pool: Max-pooling. FC: Fully connected layer.}
    \label{fig:network_architecture}
\end{figure}

To achieve a high-accuracy lightweight model suitable for rapid disease diagnosis on mobile devices, we developed architectures based on convolutional neural networks (CNN) and deep neural networks (DNN). The input data specifications for both architectures are determined by the chosen feature extraction method, with all features used as input, as illustrated in \hyperref[fig:network_architecture]{Fig.\ref{fig:network_architecture}}.

The CNN architecture consists of a single convolutional layer with a kernel size of $1 \times 1$, designed to increase the number of output channels. We set the number of output channels to 5 and selected the ReLU function as the activation function. This is followed by a $2 \times 2$ max-pooling layer and a dropout rate of 0.25 to reduce overfitting and enhance the model’s generalization capability. Subsequently, there is a fully connected layer of size $39\times1$ with a dropout rate of 0.5, followed by a sigmoid activation function to obtain the final classification result in the linear layer.

The DNN architecture includes four fully connected layers. Nonlinear transformations between each layer are performed using the ReLU activation function, and dropout rates of 0.25 are applied between the second, third, and fourth fully connected layers to exclude some data. The final classification results are obtained using a sigmoid activation function in the linear layer.

\subsection{Evaluation Metrics}
\textbf{(Cross-Validation)} To evaluate the model performance and ensure the generalizability of the results, we employed 3-fold cross-validation. The dataset was randomly split into three parts, with each part used as the validation set while the other two parts were used for training. This process was repeated three times, with each part of the dataset serving as the validation set once. The evaluation scores from all three folds were averaged to obtain the final performance metrics. This cross-validation method helps to mitigate overfitting and ensures that the model is evaluated on different subsets of the data, providing a more robust measure of its performance. Additionally, it ensures that every data point is used both for training and validation, improving the reliability of the model's performance estimation.

During this process, we did not use an additional split to create a separate validation set. Instead, one of the three parts of the dataset was designated as the validation set in each fold. This ensured that every data point was used both for training and validation, and helped provide a more robust evaluation of the model's performance while avoiding any data leakage.

Before each fold in the 3-fold cross-validation process, we first randomly shuffle the dataset to prevent potential distribution bias. For each fold, we ensure that there is no overlap between the training and test sets at the patient ID level to avoid data leakage. Additionally, PCA dimensionality reduction and RF feature selection is performed separately within each training set of each fold, preventing test set information from leaking into the feature engineering process and ensuring the independence of model evaluation.

\textbf{(Model Performance)} In addition, we use accuracy, F1 score, precision, and recall as the metrics to evaluate the performance of our model in the study. Their values are given by:
\begin{equation}
\renewcommand{\arraystretch}{2.5}
\begin{cases}
\text{Accuracy} = \frac{\text{TP} + \text{TN}}{\text{TP} + \text{FP} + \text{TN} + \text{FN}} \\
\text{F1 Score} = 2 \times \frac{\text{Precision} \times \text{Recall}}{\text{Precision} + \text{Recall}} \\
\text{Precision} = \frac{\text{TP}}{\text{TP} + \text{FP}} \\
\text{Recall} = \frac{\text{TP}}{\text{TP} + \text{FN}}
\end{cases}
\end{equation}
where TP, FN, FP, FN are given by confusion matrix.

\section{Statistical Results}
\subsection{Feature Selection Results}

Using the selected statistical metrics, and employing the feature selection techniques, \hyperref[tab:stats_results]{Tab.\ref{tab:stats_results}} demonstrates the feature extracted by RF (23 features) and PCA (27 features), respectively:

\begin{table}[h]
\centering
\caption{Results of feature extraction of random forest and principal component analysis}
\begin{tabular}{p{6cm}|p{6cm}}
\hline \hline 
\textbf{Random Forest Selected Features} & \textbf{PCA Selected Features} \\ \hline
Fundamental Frequency max Value & Fundamental Frequency max Value \\ \hline
Fundamental Frequency mean Value & Fundamental Frequency mean Value \\ \hline
Fundamental Frequency skew Value & Fundamental Frequency skew Value \\ \hline
Log Energy min value & Log Energy min value \\ \hline
Log Energy mean value & Log Energy max value \\ \hline
Log Energy std value & Log Energy mean value \\ \hline
Short-term Energy min value & Short-term Energy min value \\ \hline
Short-term Energy mean value & Short-term Energy max value \\ \hline
Short-term Energy std value & Short-term Energy range value \\ \hline
Zero Crossing Rate max value & Short-term Energy mean value \\ \hline
Zero Crossing Rate range value & Short-term Energy std value \\ \hline
Zero Crossing Rate mean value & Zero Crossing Rate min value \\ \hline
Zero Crossing Rate std value & Zero Crossing Rate max value \\ \hline
Zero Crossing Rate kurt value & Zero Crossing Rate range value \\ \hline
Sound Pressure Level min value & Zero Crossing Rate std value \\ \hline
Sound Pressure Level mean value & Zero Crossing Rate skew value \\ \hline
Sound Pressure Level std value & Zero Crossing Rate kurt value \\ \hline
MFCC min value & Sound Pressure Level min value \\ \hline
MFCC max value & Sound Pressure Level max value \\ \hline
MFCC mean value & MFCC min value \\ \hline
MFCC std value & MFCC max value \\ \hline
MFCC skew & MFCC range value \\ \hline
MFCC kurt value & MFCC mean value \\ \hline
 & MFCC std value \\ \hline
 & MFCC skew value \\ \hline
 & MFCC kurt value \\ \hline
\end{tabular}
\label{tab:stats_results}
\end{table}

\begin{figure}[h]
    \centering
    \includegraphics[width=1.0\linewidth]{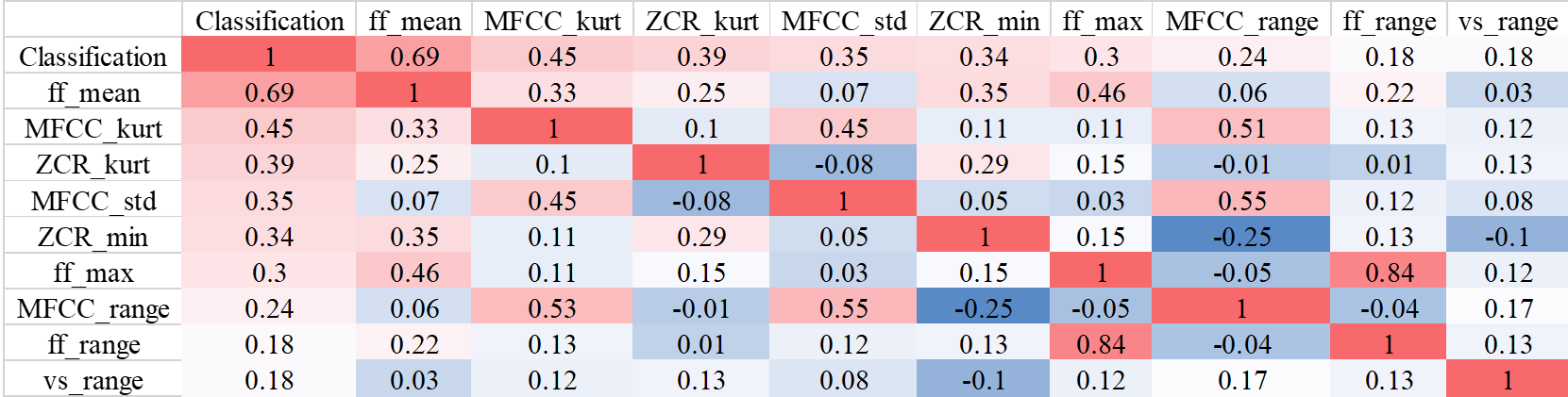}
    \caption{Heatmap of top 8 correlated coefficients from correlation analysis}
    \label{fig:heatmap}
\end{figure}

The results of top 8 correlated features  and corresponding correlation coefficients can be seen in the \hyperref[fig:heatmap]{Fig.\ref{fig:heatmap}}. In the following study we will utilize these to train lightweight deep learning-based model. The reduced dimension of the features can be seen in \hyperref[tab:dimension_reduction]{Tab.\ref{tab:dimension_reduction}}.

By leveraging these highly correlated features, we construct more robust models for effective COVID-19 detection. This strategy underscores the importance of the feature selection process in building precise and efficient diagnostic models.

Despite the rise of newer dimension reduction methods, we opted for PCA and RF because of their simplicity, efficiency, and strong performance in our dataset, which consists of nasal breath sound recordings. Moreover, the trade-off between newer techniques and the computational cost was a critical factor in our choice. As demonstrated in recent literature, PCA remains effective in preserving the most important features while significantly reducing dimensionality, thereby improving model training efficiency without sacrificing accuracy.

\begin{table}[h]
\centering
\caption{Feature Dimension Reduction Results. The values represent the dimension of features after reduction.}
\begin{tabular}{lccccccc}
\toprule
\textbf{Feature} & \textbf{Maximum} & \textbf{Minimum} & \textbf{Range} & \textbf{Std} & \textbf{Mean} & \textbf{Skew} & \textbf{Kurt} \\
\midrule
Voiced sound            & 1 & 1 & 1 & 1 & 1 & 0 & 0 \\
Unvoiced sound          & 1 & 1 & 1 & 1 & 1 & 0 & 0 \\
Effective Speech Segments & 1 & 1 & 1 & 1 & 1 & 0 & 0 \\
Fundamental Frequency   & 1 & 1 & 1 & 1 & 1 & 0 & 0 \\
Log Energy              & 1 & 1 & 1 & 1 & 1 & 1 & 1 \\
Short-term Energy       & 1 & 1 & 1 & 1 & 1 & 0 & 0 \\
Zero Crossing Rate      & 1 & 1 & 1 & 1 & 1 & 0 & 0 \\
Sound Pressure Level    & 1 & 1 & 1 & 1 & 1 & 0 & 0 \\
MFCC                    & 1 & 1 & 1 & 1 & 1 & 1 & 1 \\
\bottomrule
\end{tabular}
\label{tab:dimension_reduction}
\end{table}

\section{Experiment Results}
\begin{table}[h]
\centering
\caption{Accuracy for different models with different feature selection methods and results.}
\label{tab:accuracy_results}
\resizebox{\textwidth}{!}{%
\begin{tabular}{lcccc}
\toprule
\textbf{Model Acc} & \textbf{All 57 Features} & \textbf{RF Selected 23 Features} & \textbf{PCA Selected 27 Features} & \textbf{Correlation Selected 8 Features} \\
\midrule
CNN  & 76.55\% & 84.29\% & 86.69\% & 60.90\% \\
DNN  & 97.67\% & 96.28\% & 96.86\% & 88.31\% \\
\bottomrule
\end{tabular}%
}
\end{table}

\begin{table}[h]
\centering
\caption{F1 score for different models with different feature selection methods and results.}
\label{tab:f1score_results}
\resizebox{\textwidth}{!}{%
\begin{tabular}{lcccc}
\toprule
\textbf{Model F1} & \textbf{All 57 Features} & \textbf{RF Selected 23 Features} & \textbf{PCA Selected 27 Features} & \textbf{Correlation Selected 8 Features} \\
\midrule
CNN  & 70.53\% & 82.51\% & 86.71\% & 67.06\% \\
DNN  & 98.43\% & 93.11\% & 93.63\% & 86.91\% \\
\bottomrule
\end{tabular}%
}
\end{table}

The experimental results (\hyperref[tab:accuracy_results]{Tab.\ref{tab:accuracy_results}} and \hyperref[tab:f1score_results]{Tab.\ref{tab:f1score_results}}) demonstrated that the Deep Neural Network (DNN) model achieved the highest accuracy of 97.67\% on the full feature dataset. In comparison, the initial accuracy of the CNN model on the same dataset was 76\%. Notably, the accuracy of the CNN model improved by nearly 10\% after applying feature filtering techniques using RF and PCA. However, when the CNN model was applied to a dataset refined by correlation selection, its performance dropped by approximately 16\% compared to the full feature dataset.

In contrast, the DNN model’s accuracy only slightly decreased by about 1\% despite the reduction of nearly 30 features through RF and PCA feature filtering. The corresponding F1 score reduction was about 5\%. Even with nearly 50 features eliminated from the relevance-selected dataset, the DNN model still achieved an accuracy of 88\% and an F1 score of 86\%. This indicates that the DNN model maintained high performance across different feature subsets, demonstrating greater robustness and adaptability in managing acoustic features for COVID-19 diagnosis.

It is important to note that training models on the full feature dataset can introduce noise. This noise increases the likelihood of overfitting, where the model performs well on the training data but poorly on unseen data. The feature filtering techniques, such as RF and PCA, help mitigate this issue by removing irrelevant features, thus enhancing the model’s generalizability and reducing overfitting risks. The experimental results confirm that the DNN model is more resilient to feature subset variations, maintaining high effectiveness and reliability for COVID-19 diagnosis.

Note: Due to computational constraints and dataset limitations, visualizations of the cross-validation and performance metrics were not included in this manuscript. However, we recognize the importance of such visualizations for enhancing the clarity of model performance and plan to incorporate them in future versions of this work.

\section{Discussion}
In this study, we introduced a lightweight classification model for COVID-19 detection using nasal breathing sounds, aiming to contribute to the growing field of non-invasive diagnostics. The key accomplishments of our study, the novel contributions, and the limitations are discussed in the following texts. 

Our research successfully developed a highly accurate and efficient method for detecting COVID-19 through nasal breathing sounds recorded on smartphones. The classification model achieved an impressive accuracy rate of 97\% and an F1 score of 98\% when utilizing the full feature set. By employing DNN, our model consistently outperformed CNN, particularly with datasets containing comprehensive audio features. These results suggest the robustness and effectiveness of DNN in handling complex audio signal features, making it a viable tool for disease detection in real-world. 

The inclusion of patients with neoplastic disease in this study introduces potential biases, as their nasal breathing sound characteristics may differ from those of the general population, particularly in individuals without underlying health conditions. This could affect the model’s ability to generalize across diverse populations. In future work, we plan to explore this potential bias more thoroughly and consider methods to mitigate its impact, such as testing the model on separate datasets or applying stratified sampling techniques.

A key innovation of our study lies in the use and processing of nasal breathing sounds as input data. Unlike general breath sounds, nasal breathing sounds are more reflective of the nasal cavity, nasopharynx, and upper airway characteristics, which are crucial in detecting COVID-19-related respiratory symptoms. This makes nasal breathing sounds a more targeted and informative source of data compared to broader, less specific breath sounds. Furthermore, our approach to feature selection focused on using nine key acoustic features, with a particular emphasis on voiced sounds, which are essential for capturing the dynamics of respiratory health and disease-related changes.

The method of dimensionality reduction using Random Forest and Principal Component Analysis (PCA) enhanced the representativeness of the audio features, thus improving model performance. Although we employed PCA and RF for feature dimensionality reduction and selection, dataset bias may still exist, such as the potential omission of certain high-dimensional features. Future research could further optimize this process by incorporating additional data augmentation strategies or integrating multiple dimensionality reduction techniques, such as t-SNE or Autoencoder.

The reduction in feature set from 57 to 23 and ultimately to 8 features resulted in the maintenance of high accuracy and F1 scores, demonstrating the effectiveness of both feature selection and dimensionality reduction strategies. These results align with recent studies on acoustic signal analysis, which highlight the importance of selecting relevant features and reducing dimensionality for improving the efficiency of machine learning models in health diagnostics\cite{qureshi2024eml,brown2020exploring,coppock2024audio}.

We observed that dimensionality reduction improved the performance of the CNN, enhancing its ability to classify COVID-19 positive and negative samples. However, for the DNN, performance was actually best when the full feature set was used without any dimensionality reduction. This suggests that while dimensionality reduction is useful for some models, it may not be beneficial for others, such as the DNN, where reducing the number of features resulted in a slight drop in accuracy.

Despite the promising results, our study has certain limitations that warrant further investigation. Firstly, while our study incorporated basic noise reduction techniques, more sophisticated methods are needed to address the noise interference commonly present in acoustic recordings, particularly in uncontrolled environments. Secondly, although our dataset contains 128 samples, expanding it to include more diverse and temporally varied samples would significantly improve the model's generalizability and robustness across different populations, environments, and COVID-19 variants. Thirdly, while we focused primarily on feature extraction and classification, future research should explore other machine learning models, ensemble methods, and hybrid architectures to enhance reliability and accuracy further. Finally, the real-world clinical application of our model requires validation through prospective studies to ensure its feasibility and utility in front-line healthcare settings, where the model could aid in rapid COVID-19 detection and assist healthcare professionals in decision-making.

Beyond COVID-19, the potential of our methodology extends to diagnosing other respiratory diseases such as influenza, bronchitis, and chronic obstructive pulmonary disease (COPD). Similar to COVID-19, these diseases can cause characteristic changes in nasal breathing sounds and vocal fold vibrations, which can be captured and analyzed using similar acoustic features and machine learning models. Notably, studies such as [26] have demonstrated the utility of sound-based diagnostic methods in identifying anomalies related to respiratory diseases, showing how such models can provide early diagnosis and continuous monitoring in real-world applications. Future research should investigate the applicability of our approach across a broader spectrum of respiratory conditions, enhancing early detection, patient management, and the use of non-invasive, cost-effective methods in diverse healthcare settings. Integrating our approach with wearable devices or smartphone applications can enable continuous, real-time monitoring, significantly improving healthcare accessibility, especially in low-resource settings.

\section{Conclusion}
Our study highlights the potential of nasal breathing sounds as a reliable, non-invasive diagnostic tool for detecting diseases like COVID-19 using a novel lightweight Dense-ReLU-Dropout model. By integrating advanced feature selection techniques such as Random Forest and Principal Component Analysis, we achieved 97\% accuracy and a 98\% F1 score, demonstrating the feasibility of smartphone-based rapid disease detection. This approach has broader applicability to other respiratory illnesses and mobile health technologies, offering a cost-effective solution for real-time diagnostics. Future work should focus on expanding the dataset through multi-center collaborations to enhance generalizability, validating the model across diverse clinical settings, and addressing challenges such as noise interference in real-world environments to ensure robust performance and seamless clinical integration.

\section*{Acknowledgement}
Thanks to Hailin Ma, Han Lu, Jiayi Guo, Rui Su for their support during this study.

The authors wish to thank the anonymous referees for their thoughtful comments, which helped in the improvement of the presentation.

\section*{Credit authorship contribution statement}
\textbf{Jiayuan She}: Methodology, Experiment, Software, Writing - review \& editing. \textbf{Peiqi Li, Renxing Li}: Software, Visualization, Writing - original draft \& review \& editing. \textbf{Shengkai Li}: Software, writing - review. \textbf{Lin Shi, Ziling Dong}: Software, Validation, Data Curation, Project Management. \textbf{Liping Gu}: Conceptualization, Formal analysis, Data Curation, Visualization. \textbf{Zhao Tong, Zhuochang Yang}: Project administration, Writing - review \& editing. \textbf{Jiangang Chen, Liang Feng, Yajie Ji}: Inspiration, Conceptualization, Writing - review \& editing, Supervision.

\section*{Declaration of Interests}
The authors declare that they have no known competing financial interests or personal relationships that could have appeared to influence the work reported in this paper.

\section*{Ethics Approval}
This study was approved by the Ethics Committee of Shanghai Sixth People’s Hospital Affiliated to Shanghai Jiaotong University School of Medicine (Approval Number: 2022-KY-050(K)) with the informed consent of all participants.

\section*{Fundings}
This work was partially supported by the National Natural Science Foundation of China (Grant No. 82151318); Science and Technology Commission of Shanghai (Grant No. 22DZ2229004, 22JC1403603, 21Y11902500); the Key Research \& Development Project of Zhejiang Province (2024C03240); Scientific Development funds for Local Region from the Chinese Government in 2023 (Grant No. XZ202301YD0032C); Jilin Province science and technology development plan project (Grant No. 20230204094YY), 2022 "Chunhui Plan" cooperative scientific research project of the Ministry of Education.

\section*{Code and Data Availability}
The code and data used in this study are available upon reasonable request. Interested researchers may contact the corresponding author to obtain access.

\bibliographystyle{IEEEtran}
\bibliography{ref.bib}
\end{document}